\documentclass{article}

\usepackage{graphicx} % more modern
\usepackage{subfigure}

\usepackage{natbib,natbibspacing}

\usepackage{algorithm}
\usepackage{algorithmic}

\usepackage[bookmarks=false]{hyperref}

\usepackage{nips12submit_e,times}
\usepackage{multirow}

\usepackage{amssymb}

\newcommand{\bm}[1]{\mathrm{\mathbf{#1}}}
\def\Tr{\mathrm{T}}
\newcommand{\RR}{\mathbb{R}}
\newcommand{\HH}{\mathbb{H}}
\newcommand{\EE}{\mathbb{E}}

\newcommand\xx{\bm{x}}
\newcommand\yy{\bm{y}}
\newcommand\aaa{\bm{a}}
\newcommand\bbb{\bm{b}}
\newcommand\qqq{\bm{q}}
\newcommand\ppp{\bm{p}}
\newcommand\Pp{\bm{P}}
\newcommand\Qq{\bm{Q}}

\def\sign{\mathrm{sign}\,}

\title{Multimodal similarity-preserving hashing}

\author{Jonathan Masci$^{1,2,3}$ \\
{\tt \small jonathan@idsia.ch} \\
\and Michael M. Bronstein$^{2}$ \\
{\tt \small michael.bronstein@usi.ch}
\and Alexander A. Bronstein$^{4}$ \\
{\tt\small bron@eng.tau.ac.il}
\and J{\"u}rgen Schmidhuber$^{1,2,3}$\\
{\tt \small juergen@idsia.ch} \\
\vspace{3mm}\\
\small $^1$Istituto Dalle Molle di Studi sull'Intelligenza Artificiale (IDSIA), Manno, Switzerland\\
\small $^2$Faculty of Informatics, Universit{\`a} della Svizzera Italiana (USI), Lugano, Switzerland\\
\small $^3$Scuola Universitaria Professionale della Svizzera Italiana (SUPSI), Lugano, Switzerland\\
\small $^4$School of Electrical Engineering, Tel Aviv University, Tel Aviv, Israel.\\}

\nipsfinalcopy

\begin{document}

\maketitle

\begin{abstract}

%In this paper, 
We introduce an efficient computational framework for hashing data belonging to multiple modalities into a single representation space where they become mutually comparable.
The proposed approach is based on a novel coupled siamese neural network architecture and allows unified treatment of intra- and inter-modality similarity learning. Unlike existing cross-modality similarity learning approaches, our hashing functions are not limited to binarized linear projections and can assume arbitrarily complex forms.
We show experimentally that our method significantly outperforms state-of-the-art hashing approaches on multimedia retrieval tasks.
\end{abstract}

\section{Introduction}

%The notion of 
Similarity is a fundamental notion underlying a variety of computer vision, pattern recognition, and machine learning tasks ranging from retrieval, ranking, classification, and clustering to object detection, tracking, and registration. In all these problems, one has to quantify the degree of similarity between objects usually represented as feature vectors. While in some cases domain-specific knowledge dictates a natural similarity function, most generally a ``natural'' measure of similarity is rather elusive and cannot be constructed without side information provided e.g. through human annotation.
An even more challenging setting frequently arises in tasks involving multiple media or data coming from different modalities. 
For example, a medical image of the same organ can be obtained using different physical processes such as CT and MRI; a multimedia search engine may perform queries in a corpus consisting of audio, video, and textual information. 
While domain knowledge can be used to construct reasonable similarity functions for each data modality, it is much more challenging to create a consistent and meaningful similarity measure \emph{across} them.

\paragraph{Previous work}
The idea of constructing similarity measures suitable to specific data has been thoroughly explored by the statistics and machine learning communities.
One can roughly divide similarity learning methods into unsupervised and supervised. The former class uses only the data  with no additional side information. Unsupervised methods include PCA and its kernelized version (\cite{scholkopf1997kernel}) that approximate the data globally by their second-order statistics either in the original Euclidean space or in a feature space represented by a kernel; and various local embedding methods such as the locally linear embedding (\cite{roweis2000nonlinear}), Laplacian eigenmaps (\cite{belkin2003laplacian}), and diffusion maps (\cite{coifman2006diffusion}), which are all based on the assumptions that the data residing in a high-dimensional Euclidean space actually belong to a low-dimensional manifold, a parametrization of which is looked for. Unsupervised methods are inherently limited due to their inability to incorporate side information into the learning process.

Supervised methods can be further subdivided according to the type of side information they rely on. Class labels is the most straightforward way of specifying side information, and is used in methods dating back to LDA (\cite{johnson2002applied}) and its kernelized version (\cite{mika1999fisher}) as well as more modern approaches of 
\cite{xing2002distance,weinberger2009distance}.
Other methods accept side information in the form of knowingly similar and dissimilar pairs (\cite{davis2007information}) or triplets of the form ``$x$ is more similar to $y$ than $z$'' (\cite{shen2009positive,mcfee2009partial}). A family of methods referred to as multidimensional scaling (MDS) rely on metric dissimilarity values supplied on a training set of pairs of data vectors, and seek for a Euclidean representation reproducing them as faithfully as possible (\cite{borg2005modern}). Once the embedding into the representation space has been learned, similarity to new, unseen data is computed either directly (if the metric admits a parametric representation), or using an out-of-sample extension.

Similarity learning methods can also be classified by the type of the produced similarity functions. A significant class of practical methods learns a linear projection making the Euclidean metric optimal -- this is essentially equivalent to learning a Mahalanobis distance (\cite{weinberger2009distance,shen2009positive}). Kernelized versions of these approaches are often available in the cases where the data have an intricate structure that cannot be captured by a linear transformation.

Hashing approaches represent the data as binary codes to which the Hamming metric is subsequently applied as the measure of similarity. These methods include the family of locality sensitive hashing (LSH) (\cite{gionis1999similarity}), 
%that embeds various standard similarity measures into the Hamming metric, 
and the recently introduced spectral hashing (\cite{weiss2008spectral}). These approaches are mainly used to construct an efficient approximation of some trusted standard similarity such as the Jaccard index or the cosine distance, and are inapplicable if side information has to be relied upon. \citet{shakhnarovich2003fast} proposed to construct optimal LSH-like hashes (referred to as {\em similarity-sensitive hashing} or SSH) using supervised learning. More efficient approaches have been subsequently proposed by \citet{torralba2008small} and \citet{strecha2010ldahash}.
%with impressive performance in large-scale context based retrieval and image-based localization applications.

%Finally, kernel learning methods \cite{?} construct an inner product function instead of a metric.

The extension of similarity learning to multi-modal data has been addressed in the literature only very recently. \citet{bronstein2010data} used a supervised learning algorithm based on boosting to construct hash functions of data belonging to different modalities in a way that makes them comparable using the Hamming metric. This method can be viewed as an extension of SSH to the multimodal setting, dubbed by the authors {\em cross-modal SSH} (CM-SSH), and it enjoys the compactness of the representation and the low complexity involved in distance computation.
 \citet{mcfee2011learning} proposed to learn multi-modal similarity using ideas from multiple kernel learning (\cite{bach2004multiple,mcfee2009partial}). Multi-modal kernel learning approaches have been proposed by \citet{lee2009learning} for medical image registration, and by \citet{weston2010large}. % under the name of WSABIE (short for web-scale annotation by image embedding) for large-scale multi-modal information retrieval. %To the best of our knowledge, these are the only studies addressing multi-modal similarity learning.
The main disadvantage of the latter is % WSABIE approach is
the fact that it is limited to linear projections only.
The framework proposed by \citet{mcfee2011learning} can be kernelized, but it involves the computationally expensive semidefinite programming, which limits scalability. Also, both algorithms produce continuous Mahalanobis metrics, which is disadvantageous both in computational and storage complexity especially when dealing with large-scale data. %; this is a clear disadvantage when compact binary representation of the data is desired.
The appealing property of similarity-preserving hashing methods like the CM-SSH \cite{bronstein2010data} is the compactness of the representation and the low complexity involved in distance computation.
%bit codes produced  by embedding the data into the Hamming metric space, and the computational efficiency of the Hamming metric.

%However, CM-SSH is limited to linear projections which may not capture the structure of the data. Furthermore, it accounts only for the similarity {\em across} modalities, completely ignoring the data similarity {\em within} each modality. Finally, CM-SSH uses relaxation to solve the underlying optimization problem.

\paragraph{Contributions}

This paper is motivated by the work of \citet{bronstein2010data} on multimodal similarity-preserving hashing.
We propose a novel multi-modal similarity learning framework based on neural networks. Our approach has several advantages over the state-of-the-art.
%
%\begin{itemize}
%\item
%
First, we combine intra- and inter-modal similarity into a single framework. This allows exploiting richer information about the data and can tolerate missing modalities; the latter is especially important in sensor networks where one or more sensors may fail or in application like multimedia retrieval where it is hard to obtain reliable samples of cross-modal similarity. We show that previous works can be considered as particular cases of our model. 
%    To the best of our knowledge, ours is the first approach allowing to take advantage of both inter- and intra-modality similarities.
%
%\item
%
%Produces compact binary code representation of the data, thus reducing storage complexity and the computational complexity of the similarity function, and is better amenable for efficient indexing.
%
%\item
Second, we solve the full optimization problem without resorting to relaxations as in SSH-like methods; it has been recently shown that such a relaxation degrades the hashing performance (see e.g., \cite{strecha2010ldahash,masci2011descriptor}). 
%As a result, we obtain an optimal solution rather than a sub-optimal one.
%
Third, we introduce a novel coupled siamese neural network architecture to solve the optimization problem underlying our multi-modal hashing framework. 
Fourth, the use of neural networks can be very naturally generalized to more complex non-linear projections using multi-layered networks, thus allowing to produce  embeddings of arbitrarily high complexity. We show experimental result on several standard multimodal datasets demonstrating that our approach compares favorably to state-of-the-art algorithms.

%\end{itemize}

%The rest of the paper is organized as follows: In Section~2 we formulate the multi-modal similarity learning problem and introduce notation. In Section~3, the cross modality similarity sensitive hashing algorithm is briefly overviewed. In Section~4, we introduce the multimodal ANN hashing framework. Section~5 is dedicated to an experimental evaluation. Finally, Section~6 concludes the paper.

%\section{Multimodal similarity preserving hashing}
\section{Background}

Let $X \subseteq \RR^n$ and $Y \subseteq \RR^{n'}$ be two spaces representing data belonging to different modalities (e.g., $X$ are images and $Y$ are text descriptions).
Note that even though we assume that the data can be represented in the Euclidean space, the similarity of the data is not necessarily Euclidean and in general can be described by some metrics $d_X : X\times X \rightarrow \RR_+$ and $d_Y : Y\times Y \rightarrow \RR_+$, to which we refer as {\em intra-modal dissimilarities}.
Furthermore, we assume that there exists some {\em inter-modal dissimilarity} $d_{XY}: X\times Y \rightarrow \RR_+$ quantifying the ``distance'' between points in different modality.
The ensemble of intra- and inter-modal structures $d_X, d_Y, d_{XY}$ is not necessarily a metric in the strict sense. In order to deal with these structures in a more convenient way, we try to represent them in a common metric space. In particular, the choice of the Hamming space offers significant advantages in the compact representation of the data as binary vectors and the efficient computation of their similarity. 

%\subsection{Multimodal similarity-preserving hashing}

{\em Multimodal similarity-preserving hashing} is the problem of represent the data from different modalities $X, Y$ in a common space $\HH^m = \{ \pm 1\}^m$ of $m$-dimensional binary vectors with the Hamming metric $d_{\HH^m}(a, b) =  \frac{m}{2} - \frac{1}{2} \sum_{i=1}^m a_i b_i$ by means of two embeddings, $\xi: X \rightarrow \HH^m$ and $\eta : Y \rightarrow \HH^m$ mapping similar points as close as possible to each other and dissimilar points as distant as possible from each other, such that $d_{\HH^m} \circ (\xi \times \xi) \approx d_{X}$, $d_{\HH^m} \circ (\eta \times \eta) \approx d_{Y}$, and $d_{\HH^m} \circ (\xi \times \eta) \approx d_{XY}$.
In a sense, the embeddings act as a {\em metric coupling}, trying to construct a single metric that preserves the intra- and inter-modal similarities.
%
%
%\subsection{Cross-modality hashing problem}
%
A simplified setting of the multimodal hashing problem used in \cite{bronstein2010data} is {\em cross-modality similarity-preserving hashing}, in which only the inter-modal dissimilarity $d_{XY}$ is taken into consideration and $d_X, d_Y$ are ignored. 
To the best of our knowledge, the full multimodal case has never been addressed before.

For simplicity, in the following discussion we assume the side information given as the intra- and inter-modal dissimilarities to be binary, $d_X, d_Y, d_{XY} \in \{0, 1\}$, i.e., a pair of points can be either similar or dissimilar. This dissimilarity is usually unknown and hard to model, however, it should be possible to sample $d_X, d_Y, d_{XY}$ on some subset of the data $X' \subset X, Y' \subset Y$.
This sample can be represented as sets of similar pairs of points ({\em positives})
%
%\begin{eqnarray*}
%\mathcal{P}_X &=& \{ (x \in X', x', \in X') : d_{X}(x,x') = 0\} \\
%\mathcal{P}_Y &=& \{ (y \in Y', y'\in Y') : d_{Y}(y,y') = 0\} \\
%\mathcal{P}_{XY} &=& \{ (x \in X', y\in Y') : d_{XY}(x,y) = 0\},
%\end{eqnarray*}
%({\em positives}) 
$\mathcal{P}_X = \{ (x \in X', x', \in X') : d_{X}(x,x') = 0\}$, $\mathcal{P}_Y = \{ (y \in Y', y'\in Y') : d_{Y}(y,y') = 0\}$, and $\mathcal{P}_{XY} = \{ (x \in X', y\in Y') : d_{XY}(x,y) = 0\}$,
and likely defined sets $\mathcal{N}_X, \mathcal{N}_Y$, and $\mathcal{N}_{XY}$ of dissimilar pairs of points ({\em negatives}). 
In many practical applications such as image annotation or text-based image search, it might be hard to get the inter-modal positive and negative pairs, but easy to get the intra-modal ones. 

%\begin{eqnarray*}
%\mathcal{N}_X &=& \{ (x \in X', x'\in X') : d_{X}(x,x') = 1\} \\
%\mathcal{N}_Y &=& \{ (y \in Y', y'\in Y') : d_{Y}(y,y') = 1\} \\
%\mathcal{N}_{XY} &=& \{ (x \in X', y\in Y') : d_{XY}(x,y) = 1\}.
%\end{eqnarray*}

%$\mathcal{N}_X = \{ (x \in X', x'\in X') : d_{X}(x,x') = 1\}$,
%$\mathcal{N}_Y = \{ (y \in Y', y'\in Y') : d_{Y}(y,y') = 1\}$, and $\mathcal{N}_{XY} = \{ (x \in X', y\in Y') : d_{XY}(x,y) = 1\}$.

The problem of multimodal similarity-preserving hashing boils down to find two embeddings $\xi: X \rightarrow \HH^m$ and $\eta : Y \rightarrow \HH^m$ such that $m\, d_{\HH^m} \circ (\xi \times \eta) \approx d_{XY}$ minimizing the aggregate of false positive and false negative rates,
\begin{eqnarray}
\label{eq:mmloss}
\min_{\xi,\eta}  &&  \EE \{ d_{\HH^m} \circ (\xi \times \xi) | \mathcal {P}_X \} + \EE \{ d_{\HH^m} \circ (\eta \times \eta) | \mathcal {P}_Y \} + 
\EE \{ d_{\HH^m} \circ (\xi \times \eta) | \mathcal {P}_{XY} \} - \nonumber\\ 
&& \EE \{ d_{\HH^m} \circ (\xi \times \xi) | \mathcal {N}_X \} - 
 \EE \{ d_{\HH^m} \circ (\eta \times \eta) | \mathcal {N}_Y \} - \EE \{ d_{\HH^m} \circ (\xi \times \eta) | \mathcal {N}_{XY} \}.
\end{eqnarray}

\paragraph{Cross-modality similarity sensitive hashing}

\citet{bronstein2010data} studied the particular case of cross-modal similarity hashing (without incorporating intra-modality similarity), with linear embeddings of the form $\xi(\xx) = \sign(\Pp \xx + \aaa)$ and $\eta(\yy) = \sign(\Qq \yy + \bbb)$.
Their CM-SSH algorithm constructs the dimensions of $\xi$ and $\eta$ one-by-one using boosting.
At each iteration, one-dimensional embeddings $\xi_i(\xx) = \sign(\ppp_i \xx + a_i)$ and $\eta_i(\yy) = \sign(\qqq_i \yy + b_i)$ are found using a two-stage scheme: first, the embeddings are linearized as $\xi_i(\xx) \approx \ppp_i \xx$ and $\eta_i(\yy) \approx \qqq_i \yy$ and the resulting objective is minimized to find the projection
\begin{eqnarray}
\min_{\ppp_i,\qqq_i} \EE \{ \xx^\Tr \ppp_i^\Tr \qqq_i \yy | \mathcal{P}_{XY} \} - \EE \{ \xx^\Tr \ppp_i^\Tr \qqq_i \yy | \mathcal{N}_{XY} \},
\end{eqnarray}
(here $\ppp_i$ and $\qqq_i$ are unit vectors representing the $i$th row of the matrices $\Pp$ and $\Qq$, respectively, and the expectations are weighted by per-sample weights adjusted by the boosting).
With such an approximation, the optimal projection directions $\ppp$ and $\qqq$ have a closed-form expressions using SVD of the positive and negative covariance matrices.
At the second stage, the thresholds $a_i$ and $b_i$ are found by one-dimensional search. %solved for without resorting to approximations.
%

%SSH-type approaches (and consequently, CM-SSH) have several drawbacks.
This approach has several drawbacks. 
First, CM-SSH solves a particular setting of problem~(\ref{eq:mmloss}) with $\mathcal{P}_{XY}, \mathcal{N}_{XY}$ only, thus ignoring the intra-modality similarities.
Second, the assumption of separability (treating each dimension separately) and the linearization of the objective replace the original problem with a relaxed version, whose optimization produces suboptimal solutions that tend to increase the hash sizes (or alternatively, for a fixed hash length $m$, the method manifests inferior performance; see \citet{masci2011descriptor}).
Finally, this approximation is limited to a relatively narrow class of linear embeddings that often do not capture well the structure of the data.

%In what follows, we present a framework based on a novel neural network architecture that addresses the most general case of multimodal similarity-preserving hashing, and is free %of the above disadvantages.

%While being computationally cheap, the CM-SSH approach has several disadvantages.
%This approach has several drawbacks.
%First, it does not take into account the intra-modality similarities. Second, the approximation introduced by boosting and further approximation by the linearization of the cost function give suboptimal solutions that tend to increase the hash sizes. Finally, this approximation is limited to a relatively narrow class of embeddings that often do not capture well the structure of the data
%%
%In what follows, we present a framework based on a novel neural network architecture that addresses the most general case of multimodal similarity-preserving hashing, and is free of the above disadvantages.

\section{Multimodal NN hashing}
%The main disadvantage of CM-SSH is that it solves a relaxation of the minimization problem, which is inherently sub-optimal. It also finds only linear projections which in many cases are not enough to model complex mappings and does so without taking into account the potential inter-modal similarity information.

%This approach has several drawbacks. 
%First, it does not take into account the intra-modality similarities. Second, the approximation introduced by boosting and further approximation by the linearization of the cost function give suboptimal solutions that tend to increase the hash sizes. Finally, this approximation is limited to a relatively narrow class of embeddings that often do not capture well the structure of the data
%%
%In what follows, we present a framework based on a novel neural network architecture that addresses the most general case of multimodal similarity-preserving hashing, and is free of the above disadvantages.

%\paragraph{Siamese architecture}
Our approach for multimodal hashing is related to supervised methods for dimensionality reduction and in particular extends the framework of~(\cite{SchmidhuberPrelinger:93,hadsell-chopra-lecun-06,Taylor11}),  also known as the {\em siamese architecture}. 
These methods learn a mapping onto a usually low-dimensional feature space such that similar observations are mapped to nearby points in the new manifold and dissimilar observations are pulled apart. 
%
%In its original formulation the siamese framework minimizes a loss function as the one in eq~\ref{eq:mmnnhash-loss-X}. 
In our simplest setting, the linear embedding $\xi = \mathrm{sign}(\Pp \xx + \aaa)$ is realized as a neural network with a single layer (where $\Pp$ represent the linear weights and $\aaa$ is the bias) and a sign activation function (in practice, we use a smooth approximation $\mathrm{sign}(x) \approx \mathrm{tanh}(\beta x)$). 
The parameters of the embedding can be learned using the back-propagation algorithm (\cite{Werbos:74}) minimizing the loss 
\begin{eqnarray}
\label{eq:mmnnhash-loss-X}
\mathcal{L}_{X}  & = & \frac{1}{2} \sum_{(\xx,\xx') \in \mathcal{P}_{X}} \| {\xi}(\xx) - {\xi}(\xx') \|_{2}^{2}  
			      +  \frac{1}{2} \sum_{(\xx,\xx') \in \mathcal{N}_{X}} \max\{0, m_{X} - \| {\xi}(\xx) - {\xi}(\xx') \|_{2} \}^{2}  
\end{eqnarray}
w.r.t. the network parameters $(\Pp, \aaa)$. 
%Same way, embedding $\eta$ is learned by minimizing the loss $\mathcal{L}_{Y}$ w.r.t. parameters $(\Qq,\bbb)$.  
Note that for binary vectors (when $\beta = \infty$), the squared Euclidean distance in~(\ref{eq:mmnnhash-loss-X}) is equivalent up to constants to the Hamming distance. 
The second term in~(\ref{eq:mmnnhash-loss-X}) is a {\em hinge-loss} providing robustness to outliers and produces a mapping for which negatives are pulled $m_X$ apart.   
%
%This allows to easily extend the model to incorporate more levels of non-linearity or more complex mapping.
The system is fed with pairs of samples which share the same parametrization and for which a corresponding dissimilarity is known, $0$ for positives and $1$ for negatives (thus the name {\em siamese network}, e.g. two inputs and a common output vector).
%This approach has been also successfully applied by \cite{Taylor11} to problems such as  matching %people in similar pose. % and which exhibits invariance to identity, clothing, background, lighting, shift and scale.

\begin{figure}[!h]
\begin{center}
\includegraphics[width=0.65\linewidth]{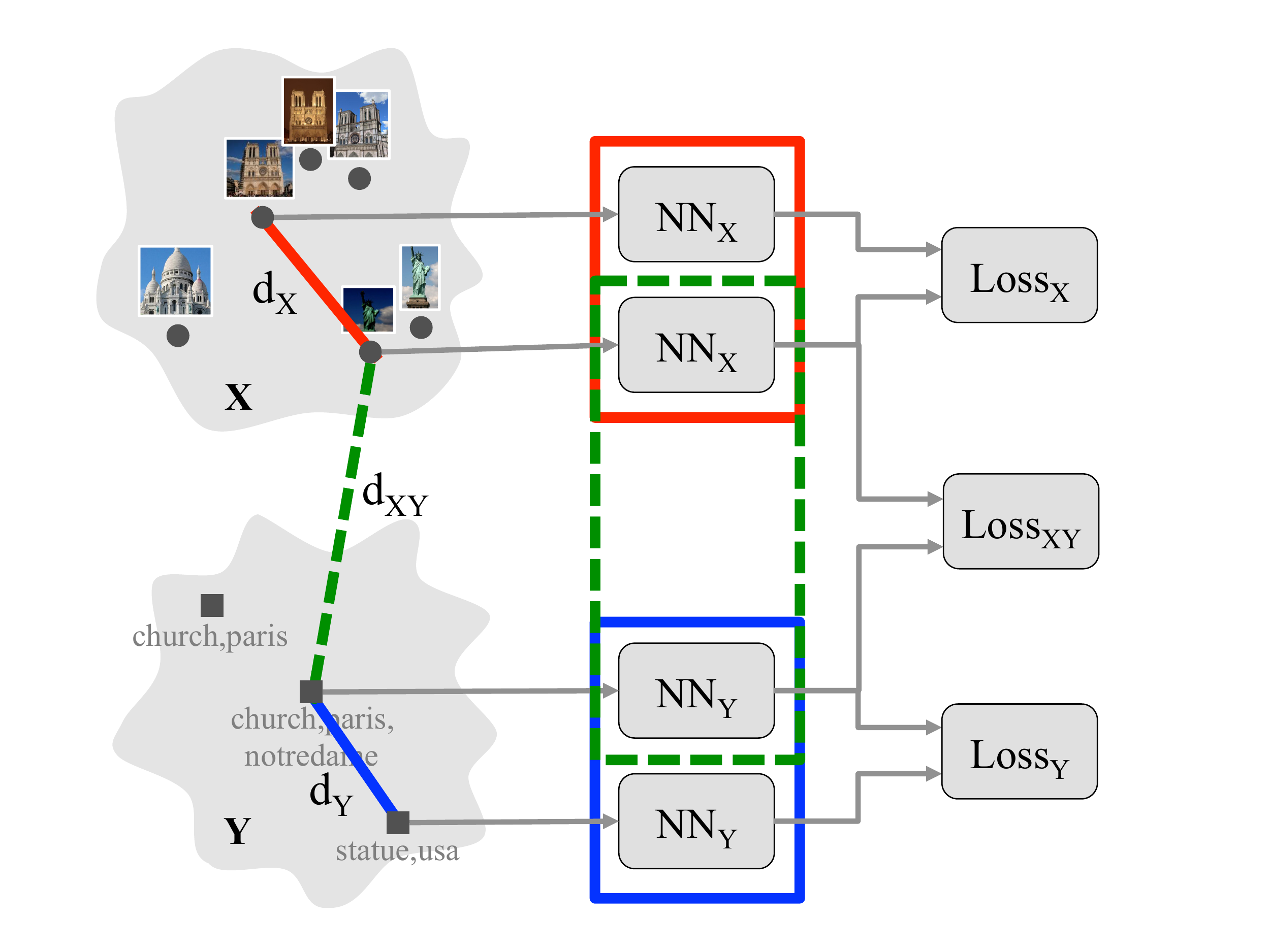}\vspace{-5mm}
\end{center}
\caption{Schematic representation of our coupled siamese network framework. Two pairs of nets, one for modality $X$ and one for modality $Y$ are coupled together by a cross-modal loss. The system learns two set of parameters.}
\label{fig:ann}
\end{figure}

\paragraph{Coupled siamese architecture} 
%
%A way of solving the ``true'' non-linear optimization is by formulating it in the neural network (ANN) framework, which to us seems a very powerful and flexible approach to tackle the problem; which also allows for exploiting numerous optimization techniques and heuristics developed in the field.
%
In the multimodal setting, we have two embeddings $\xi$ and $\eta$, each cast as a siamese network with parameters $(\Pp,\aaa)$ and $(\Qq,\bbb)$, respectively. 
%
%
%We map each modality into a {\em siamese network} \cite{SchmidhuberPrelinger:93,hadsell-chopra-lecun-06} which learns a mapping by minimizing
%%~(\ref{eq:mmnnhash-loss-X}) or~(\ref{eq:mmnnhash-loss-Y}). 
%\begin{eqnarray}
%\label{eq:mmnnhash-loss-X}
%\mathcal{L}_{X}  & = & \frac{1}{2} \sum_{(\xx,\xx') \in \mathcal{P}_{X}} \| {\xi}(\xx) - {\xi}(\xx') \|_{2}^{2}  \\  \nonumber
%			    &  + & \frac{1}{2} \sum_{(\xx,\xx') \in \mathcal{N}_{X}} \max\{0, m_{X} - \| {\xi}(\xx) - {\xi}(\xx') \|_{2}\}^{2}; 
%\end{eqnarray}
%\begin{eqnarray}
%\label{eq:mmnnhash-loss-Y}
%\mathcal{L}_{Y}  & = & \frac{1}{2} \sum_{(\yy,\yy') \in \mathcal{P}_{Y}} \| {\eta}(\yy) - {\eta}(\yy') \|_{2}^{2} \\  \nonumber
%			   &  + & \frac{1}{2} \sum_{(\yy,\yy') \in \mathcal{N}_{Y}} \max\{0, m_{Y} - \| {\eta}(\yy) - {\eta}(\yy') \|_{2}\}^{2}
%\end{eqnarray}
%w.r.t. parameters $\theta_\xi, \theta_\eta$ of the nets $\xi, \eta$. In our experiment, in order to make a fair comparison with the other approaches, we used a single layer fully connected network and hence the parameters to optimize match $\theta_\xi = \Pp,\aaa; \theta_\eta = \Qq, \bbb$. That is we learn a projection matrix and an offset vector for each of the modality.
%%
%The hinge-loss (second term in~(\ref{eq:mmnnhash-loss-X}) and~(\ref{eq:mmnnhash-loss-Y})) provides robustness to outliers and produces a mapping for which patterns are pulled at the specified margin. 
%
Such an architecture allows to learn similarity-sensitive hashing for each modality independently by minimizing the loss functions $\mathcal{L}_X, \mathcal{L}_Y$. %, by performing back propagation with respect to . 
In order to incorporate inter-modal similarity, we couple the two siamese networks by the cross-modal loss 
\begin{eqnarray}
\label{eq:mmnnhash-loss-CM}
\mathcal{L}_{XY}  & = & \frac{1}{2} \sum_{(\xx,\yy) \in \mathcal{P}_{XY}} \| {\xi}(\xx) - {\eta}(\yy) \|_{2}^{2}  
			      +  \frac{1}{2} \sum_{(\xx,\yy) \in \mathcal{N}_{XY}} \max\{0, m_{XY} - \| {\xi}(\xx) - {\eta}(\yy) \|_{2} \}^{2},
\end{eqnarray}
 thus jointly learning two sets of parameters for each modality. We refer to this model, which generalizes the siamese framework, as {\em coupled siamese networks} for which a schematic representation is shown Figure~\ref{fig:ann}.

%{\bf Multimodal hashing optimization problem. } In the general multi-modal setting both intra- and inter-modal similarities have to be considered. We propose to solve this problem in our ANN framework with the following minimization problem:

%The full loss in the coupled siamese network is given by 
%
Our implementation differs from the original architecture of \citet{hadsell-chopra-lecun-06} in the choice of the output activation function (we use $\tanh$ activation that encourages binary representations rather than a linear output layer). 
This way the maximum distance is bounded by $\sqrt{4 m}$ and by simply enlarging the margin between dissimilar pairs we enforce the learning of codes which differ by the sign of their components.
%The only way of pulling two dissimilar vectors at a given minimum distance is to have them with different sign and enlarge the magnitude of the respective values, which is bounded to $[-1, 1]$. 
Once the model is learned, hashes are produced by thresholding the output. %of the network output.

The reader should also note that the hamming distance is equal to the squared euclidean distance. Hence the loss function in eq \ref{eq:mmnnhash-loss}, when $\beta \to +\infty$, margins $=0$ and $\alpha = 1$ coincides with eq \ref{eq:mmloss}. However for optimization reasons a margin needs to be added.
%
%This architecture can be extended to arbitrarily complex mappings by adding multiple layers of non-linearities. This has the advantage of scaling linearly with the number of activations which is a very desirable property in large scale problems.

\paragraph{Training} 

The training of our coupled siamese network is performed by minimizing %e the full loss of the form 
\begin{eqnarray}
\label{eq:mmnnhash-loss}
\min_{\Pp,\aaa, \Qq, \bbb} \mathcal{L}_{XY} + \alpha_{X} \mathcal{L}_{X} + \alpha_{Y} \mathcal{L}_{Y}, 
\end{eqnarray}
where $\alpha_X, \alpha_Y$ are weights determining the relative importance of each modality. 
The loss~(\ref{eq:mmnnhash-loss}) can be considered as a generalization of the loss in~(\ref{eq:mmloss}), which is obtained by setting $\alpha_X = \alpha_Y = 1$, margins = 0, and $\beta = \infty$. 
Furthermore, setting $\alpha_{X} = \alpha_{Y} = 0$, we obtain the particular setting of cross-modal loss, whose relaxed version is minimized by the CM-SSH algorithm of \cite{bronstein2010data}.
It is also worth repeating that in many practical cases, it is very hard to obtain reliable cross-modal training samples ($\mathcal{P}_{XY}, \mathcal{N}_{XY}$) but much easier to obtain intra-modal samples ($\mathcal{P}_{X}, \mathcal{N}_{X}, \mathcal{P}_{Y}, \mathcal{N}_{Y}$). 
In the full multimodal setting ($\alpha_{X}, \alpha_{Y} > 0$), the terms $ \mathcal{L}_{X},  \mathcal{L}_{Y}$ can be considered as a {\em regularization}, preventing the algorithm from over fitting. 

We apply the back-propagation algorithm
(\cite{Werbos:74,LeCun:85,Rumelhart1986})
to get the gradient of our model w.r.t. the embedding parameters. %, that is the partial derivative of the output w.r.t. the parameters.
%
%The gradients of the loss functions $\mathcal{L}_{X}, \mathcal{L}_{Y}$ are standard and we refer the reader to~\cite{hadsell-chopra-lecun-06} for derivation.
%
The gradients of the intra- and inter-modal loss functions w.r.t. to the parameters of $\xi$ are given by \vspace{-5mm}

\begingroup
%\everymath{\scriptstyle}
%\scriptsize
%
%
\begin{eqnarray*}
\label{eq:mmnnhash-loss-derivS}
\nabla \mathcal{L}_{X} &=&  \left\{
          \begin{array}{cc}
               (\xi(\xx) - \xi(\xx'))(\nabla \xi(\xx) - \nabla \xi(\xx'))    & (\xx,\xx')\in \mathcal{P}_{X}  \vspace{2mm} \\
               %
               %(\xi(\xx) - \xi(\xx') - m_{X})(\nabla \xi(\xx) - \nabla \xi(\xx'))    
                                              \begin{array}{cc}
         	     (\xi(\xx) - \xi(\xx') - m_{X}) 
	             (\nabla \xi(\xx) - \nabla \xi(\xx')) 
               \end{array}%\vspace{2mm}
               & %(\xx,\yy)\in \mathcal{N}_{XY}   \\
                               \begin{array}{cc}
         	      (\xx,\xx')\in \mathcal{N}_{X}\,\,\,\mathrm{and} \\
	             m_{X} > \|\xi(\xx) - \xi(\xx')\|_2
               \end{array}\vspace{2mm}\\
               0 & \mathrm{else}\
          \end{array}
          \right.\\
\nabla \mathcal{L}_{XY} &=&  \left\{
          \begin{array}{cc}
               (\xi(\xx) - \eta(\yy))\nabla \xi(\xx)    & (\xx,\yy)\in \mathcal{P}_{XY}  \vspace{2mm} \\
               (\xi(\xx) - \eta(\yy) - m_{XY})\nabla \xi(\xx)    & %(\xx,\yy)\in \mathcal{N}_{XY}   \\
                               \begin{array}{cc}
         	      (\xx,\yy)\in \mathcal{N}_{XY}\,\,\,\mathrm{and} \,\,\,
	             m_{XY} > \|\xi(\xx) - \eta(\yy)\|_2
               \end{array}\vspace{2mm}\\
               0 & \mathrm{else}\
          \end{array}
          \right.
\end{eqnarray*}
\endgroup
%
%$\nabla \mathcal{L}_{XY} = \partial{\mathcal{L}_{XY}} / {\partial (\Pp,\aaa, \Qq, \bbb)} $
%|\mathcal{P} =  (\xi(\xx) - \eta(\yy)) \frac{\partial{\xi(\xx)}}{\partial{\theta_\xi}}
%\end{eqnarray}
%%
%\begin{eqnarray}
%\label{eq:mmnnhash-loss-derivD}
%\frac{\partial{\mathcal{L}_{XY}}}{\partial \theta_\xi}|\mathcal{N} =  -(m_{XY} - \|\xi(\xx) - \eta(\yy)\|_2) \frac{\partial{\xi(\xx)}}{\partial{\theta_\xi}}
%\end{eqnarray}
%with the special case of $\frac{\partial{\mathcal{L}_{XY}}}{\partial \theta_\xi}|\mathcal{N} =  0$ in case $m_{XY} > \|\xi(\xx) - \eta(\yy)\|_2$.
%
where the term $\nabla \xi = \partial{\xi} / {\partial (\Pp,\aaa)}$  %$\frac{\partial{\mathcal{L}_{XY}}}{\partial \theta_\xi}$ 
%$\nabla \xi = \partial{\xi} / {\partial (\Pp,\aaa, \Qq, \bbb)} $ and $\nabla \eta = \partial{\eta} / {\partial (\Pp,\aaa, \Qq, \bbb)} $
is the usual back-propagation step of a neural network.  
Equivalent derivation is done for the parameters of $\eta$.
%
%A schematic representation of the proposed architecture is depicted in fig.\ref{fig:ann}. We have a coupled network composed of two siamese, each one modeling a single modality. 
The model can be easily learnt jointly using any gradient-based technique such as conjugate gradient or stochastic gradient descent.
%
%Setting $\alpha_{X} = \alpha_{Y} = 0$ we obtain the particular setting of cross-modal hashing (hereinafter referred to as CM-NNhash), whereas otherwise we have a multi-modal hashing (MM-NNhash).

%Our implementation
%% for single-modality mapping 
%also differs from the original architecture of \citet{hadsell-chopra-lecun-06} in the choice of the output activation function. We do not use a linear output layer but instead a $\tanh$ activation which encourages binary representations. The only way of pulling two dissimilar vectors at a given minimum distance is to have them with different sign and enlarge the magnitude of the respective values, which is bounded to $[-1, 1]$. Once the model is learnt hashes are produced by simple thresholding of the network's output.
%
%The reader should also note that the hamming distance is equal to the squared euclidean distance. Hence the loss function in eq \ref{eq:mmnnhash-loss}, when $\beta \to +\infty$, margins $=0$ and $\alpha = 1$ coincides with eq \ref{eq:mmloss}. However for optimization reasons a margin needs to be added.
%
%This architecture can be extended to arbitrarily complex mappings by adding multiple layers of non-linearities. This has the advantage of scaling linearly with the number of activations which is a very desirable property in large scale problems.

\paragraph{Non-linear embeddings} 
Our model straightforwardly generalizes to non-linear embeddings using multi-layered network architecture. The proposed framework is in fact general and any class of neural networks can be applied to arbitrarily increase the complexity of the embedding. Deep and hierarchical models are able to model highly non-linear embeddings and scale well to large-scale data by means of fully online learning, where the parameters are updated after every input tuple presentation. This allows to sample a possibly huge space with constant memory requirements.

\section{Results}

\begin{figure*}[t!]
\begin{center}
\includegraphics[width=\linewidth]{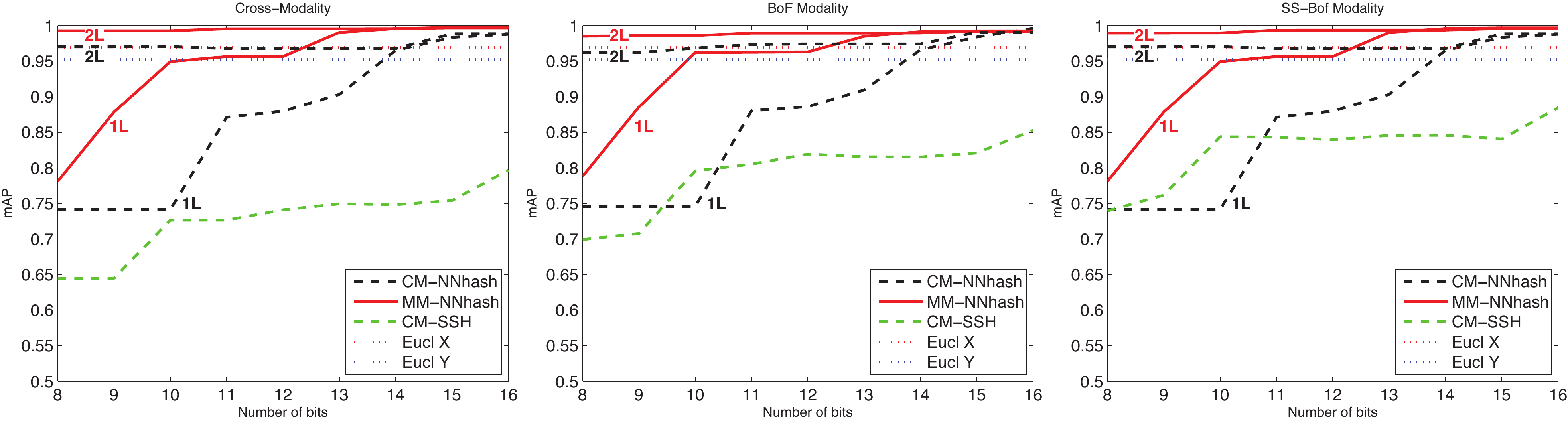} \vspace{-10mm}\\
\end{center}
\caption{Mean average precision (mAP) vs. hash length $m$ for the ShapeGoogle retrieval experiment. \textit{Left}: cross-modal (BoF--SS-BoF); 
%(matching $\xi$(BoF) and $\eta$(SS-BoF)), 
\textit{center}: BoF; % (matching $\xi$(BoF) and $\xi$(BoF')) and 
\textit{right}: SS-BoF. %(matching $\eta$(SS-BoF) and $\eta$(SS-BoF')). 
L1 and L2 refer to single and two-layer neural networks, respectively. 
Performance of raw descriptors in each modality with $L_2$ distance is shown in dotted.}
\label{fig:shapegoogle} 
\end{figure*}

%To test the performance of the proposed algorithm, we 
%used two standard datasets: Shapegoogle \cite{} and NUS \cite{}. 

We tested our algorithm on cross-modal data retrieval tasks using standard datasets from the shape retrieval and multimedia retrieval communities. 
We compared three algorithms: our coupled siamese framework in the full multimodal setting (MM-NN) and its reduced version (CM-NN), as well as CM-SSH. % \cite{bronstein2010data} . 
The single-layer version (denoted L1) of CM-NN and MM-NN realizes a linear embedding function and compares directly with CM-SSH. Two-layered version (L2) allows to obtain more complex non-linear embeddings.
%Unless written otherwise, neural networks used a single layer of the same architecture as CM-SSH, to make the comparison fair. 
%
For training the neural networks, we used conjugate gradients. %\cite{minFunc}. 
% and we run until convergence and we used the Matlab$\textsuperscript{\textregistered}$ 
%parallel computing toolbox~\cite{dowehaveto?} to make use of GPUs. This delivers a speedup of $\approx$7$\times$ with almost no code rewriting.
%
%
%
The hash functions learned by each of the methods were applied to the data in the datasets, and the Hamming distance was used to rank the matches. 
%
%Retrieval performance was evaluated using {\em precision} $P(r)$, defined as the percentage of relevant results in the first $r$ top-ranked retrieved matches. 
%
Retrieval performance was evaluated using {\em mean average precision} $mAP = \sum_{r=1}^R P(r) \cdot rel(r)$, where $rel(r)$ is the relevance of a given rank (one if relevant and zero otherwise), $R$ is the number of retrieved results, and $P(r)$ is {\em precision at $r$}, defined as the percentage of relevant results in the first $r$ top-ranked retrieved matches. %was used as a single measure of performance.
% Intuitively, mAP is interpreted as the area below the precision-recall curve.
%%
%Ideal performance retrieval performance results in first relevant match with mAP=100\%.

%
%In all our experiments we used conjugate gradient methods as optimisation algorithm~\cite{minFunc} and we run until convergence and we used the Matlab$\textsuperscript{\textregistered}$ parallel computing toolbox~\cite{dowehaveto?} to make use of GPUs. This delivers a speedup of $\approx$7$\times$ with almost no code rewriting.

%\subsection{Shapegoogle}

\paragraph{ShapeGoogle} 
In the first experiment, we reproduced the multimodal shape retrieval experiment of \citet{bronstein2010data} using the ShapeGoogle dataset \cite{bronstein2011shape}, containing 583 geometric shapes of 12 different classes subjected to synthetic transformations %(deformations, topological noise, subsampling, etc) 
as well as 456 unrelated shapes %of different objects 
(``distractors''). The goal was to correctly match a transformed shape from the query set to the rest of the dataset. 
The shapes were represented using  32-dimensional bag of geometric features (BoF) 
%obtained using dense heat kernel signature (HKS) descriptors 
%and $8\times 8$ (
and 64-dimensional spatially-sensitive bags of geometric features (SS-BoF). %, obtained using spatial co-occurrences of HKS descriptors. 
%For additional details on the dataset and the descriptors, see \cite{bronstein2010data,bronstein2011shape}.
%
%
%
To learn the hashing functions, we used positive and negative sets of size $|\mathcal{P}| = 10^4$ and  $|\mathcal{N}| = 5\times 10^4$, respectively. 
%
%Positive set consisted of different transformations of a shape from the same class; negative set consisted of shapes from different classes. 
%
%
%Hashes of length up to $m = 32$ were produced; shorter hashes were obtained by taking the first dimensions.
For CM-SSH, we used the code with settings provided by \citet{bronstein2010data}. 
For MM-NN, we used single-layer architecture with margins $m_X = m_Y = 1$, $m_{XY} = 3$ and $\alpha_{X} = 0.1$, $\alpha_{Y} = 0.3$ which we empirically found to be the best combination (additional results with different parameters are shown in Table~\ref{tab:shapegoogle-params} and in supplementary materials). 
In addition, we also show a two-layer architecture with $128$ hidden nodes. For CM-NN we used $m_{XY} = 3$ as for the single layer case. MM-NN used $m_X = m_Y = 1$, $m_{XY} = 3$ and $\alpha_{X} = 0.3$, $\alpha_{Y} = 0.3$.
%The method is however not very sensitive to the choice of these parameters.

%For this experiment we generate 10k positive and 50k negative couples. As both the modality are histogram-like representation we normalise them to have unit length.
%Both models have a margin $m = 5$ and hashes of $32$ bit are produced. 

Figure~\ref{fig:shapegoogle} and Table~\ref{tab:shapegoogle-p-at-k} shows the performance of different methods as function of hash length $m$.  
%
%We can make a few observations. 
First, we can see that NN-based methods (CM-NN and MM-NN) dramatically outperform the boosting-based CM-SSH for a fixed hash length. MM-NN achieves almost perfect performance using only $12$ bits (for comparison, CM-SSH requires almost $100$ bits to achieve similar performance). 
The reason is likely to be the fact that CM-SSH resorts to relaxation of the problem thus producing a suboptimal solution, while NN-hash solved the ``true'' optimization problem. % rather than a relaxed version thereof. 
%
%Adding multi-modal information in MM-NNhash the performance further increases as a result of the intra-modal similarities and this is particularly evident at short hash lengths.
%
Secondly, adding another layer to the neural network we obtain a non-linear hashing function, which performs dramatically better than a single-layered architecture, achieving near-perfect performance with $8$ bits. 
Thirdly, fully multimodal method (MM-NN) consistently outperforms the cross-modal version (CM-NN). We attribute this fact to the use of the intra-modal losses, acting as regularization. %usefulness of both intra- and inter-modal similarity. 

\begin{table}[htdp]
\begin{center}
\begin{minipage}[t]{0.57\linewidth}
\centering 

\begin{tabular}{rrccc}
%& \multicolumn{3}{c}{\bf mAP}  \\
%\hline
%			&  $\xi \times \eta$ & $\xi \times \xi$ & $\eta \times \eta$   \\
			&& {\bf BoF--SS-BoF} & {\bf BoF} & {\bf SS-BoF} \\
\hline
\multicolumn{2}{r}{\bf Raw}	 	& -- & 96.97\% & 95.29\%   \\
\hline
 \multirow{2}{*}{\bf MM-NN }  &  {\bf L1} & 95.63\%  &  96.29\% & 95.63\%  \\
 & {\bf L2}   & {\bf 99.56}\%  & {\bf 98.94}\% & {\bf 99.37}\%  \\
 \hline
 \multirow{2}{*}{\bf CM-NN } & {\bf L1}    & 87.97\%         & 88.60\% & 87.97\% \\
 & {\bf  L2}  & 96.77\%         & 97.40\% & 96.77\% \\
\hline
\multicolumn{2}{r}{\bf CM-SSH}	      &  74.09\%         & 81.93\%     &  83.95\%   \\	
\hline
\end{tabular}

\vspace{1mm}$12$-bit hash

\end{minipage} \hspace{1mm}
\begin{minipage}[t]{0.4\linewidth}
\centering 

\begin{tabular}{ccc}
%& \multicolumn{3}{c}{\bf mAP}  \\
%\hline
%			&  $\xi \times \eta$ & $\xi \times \xi$ & $\eta \times \eta$   \\
{\bf BoF--SS-BoF} & {\bf BoF} & {\bf SS-BoF} \\
\hline
-- & 96.97\% & 95.29\%   \\

\hline
99.69\%  & 99.18\% & {\bf 99.69}\%  \\
{\bf 99.76}\% & 99.26\% & 99.57\% \\
\hline
98.77\%         & \bf{99.64}\% & 98.99\% \\
98.83\%         & 99.08\% & 98.83\% \\
\hline
79.68\%         & 85.29\%     &  88.45\%   \\	
\hline
\end{tabular}

\vspace{1mm}$16$-bit hash

\end{minipage}

\end{center}\vspace{-3mm}
\caption{\label{tab:shapegoogle-p-at-k} Performance of different methods (mAP) on the ShapeGoogle retrieval experiment. %Hash of length $m=16$ is used. 
L1 and L2 refer to neural networks with 1 or 2 layers. }
\end{table}

\begin{figure*}[!h]
\begin{center}
\includegraphics[width=\linewidth]{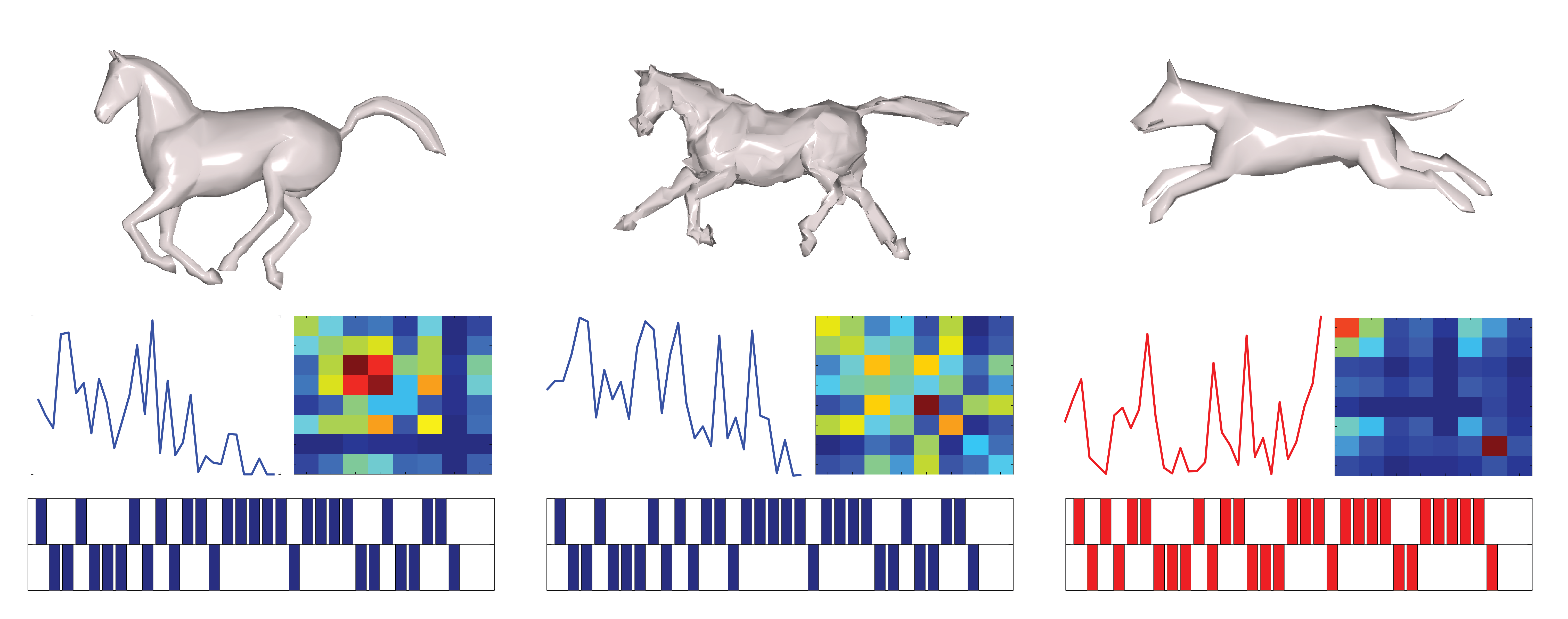}
\end{center}
\caption{ShapeGoogle retrieval example. \textit{Top}: original shapes, \textit{middle}: BoF descriptor (left) and SS-BoF (right), \textit{bottom}: MM-NNhash binary descriptors.}
\label{fig:retrieval-shapegoogle}
\end{figure*}

Figure~\ref{fig:retrieval-shapegoogle} visually exemplifies a retrieval experiment for MM-NNhash where the query shape on the left-most side is compared with a similar shape, middle, and a dissimilar one, right-most. The produced hash vectors on the bottom row are shown along with the BoF and the SS--BoF descriptors.

%\paragraph{Importance of intra-modal regularization}  
%
%%
%Second, fully multimodal method (MM-NN) outperforms the cross-modal version (CM-NN). We attribute this fact to the importance of the intra-modal regularization. %usefulness of both intra- and inter-modal similarity. 

{\bf Importance of  intra-modal regularization} is exemplified in Table~\ref{tab:shapegoogle-p-at-k-missingdata}. In this experiment, we performed training of a 1 layer net using a subset of the cross-modal data ($\mathcal{P}_{XY}, \mathcal{N}_{XY}$), while keeping the intra-modal data in the MM-NN. 
The CM-NN method manifested significant performance drop (attributed most likely to overfitting), while the performance of MM-NN remains practically unchanged.

\begin{table}[htdp]
\begin{center}
\begin{tabular}{rcccccc}
%& \multicolumn{3}{c}{\bf mAP}  \\
%\hline
%			&  $\xi \times \eta$ & $\xi \times \xi$ & $\eta \times \eta$   \\
			& \multicolumn{2}{c}{\bf BoF--SS-BoF} & \multicolumn{2}{c}{\bf BoF} & \multicolumn{2}{c}{\bf SS-BoF} \\
\hline
 	& $1/2$ & $1/10$  	& $1/2$ & $1/10$ &  $1/2$ & $1/10$ \\
\hline

%{\bf $L_2$ BoF}	 	& & & 96.97\% & 96.97\% & &     \\
%{\bf $L_2$ SS-BoF}	 & & & & & 95.29\%  & 95.29\%      \\
%\hline

%{\bf MM-NN}     & {\bf 99.02}\% & {\bf 98.13}\%  	   & {\bf 99.58}\% & {\bf 99.41}\%     & {\bf 99.02}\% & {\bf 98.13}\%  \\
%{\bf CM-NN}     & 92.24\%  & 92.95\%    &  93.25\% & 91.10\%    & 93.80\% & 92.65\%   \\

{\bf MM-NN}     & {\bf 99.53}\% & {\bf 99.28}\%  	   & {\bf 99.67}\% &  {\bf 99.63}\%     & {\bf 99.53}\% & {\bf 99.26}\% \\
{\bf CM-NN}     & 98.44\% & 95.03\% &  98.96\% & 94.35\%     & 98.51\%  & 95.83\%   \\

%\hline
%{\bf CM-SSH}	      &  \%     & 67.67\%     &  84.18\%     & 77.54\% &  & 80.79\% \\	
\hline
\end{tabular}
\end{center}\vspace{-3mm}
\caption{\label{tab:shapegoogle-p-at-k-missingdata} Performance of different methods (mAP) on the ShapeGoogle retrieval experiment where only a subset of the cross-modal correspondences is kept . 
%Hash of length $m=12$ is used and the best configuration is found through grid search.}
Hash of length $m=16$ is used and the best configuration is found through grid search.}
\end{table}

%\begin{table}[htdp]
%\begin{center}
%\begin{tabular}{rccccccccc}
%%& \multicolumn{3}{c}{\bf mAP}  \\
%%\hline
%%			&  $\xi \times \eta$ & $\xi \times \xi$ & $\eta \times \eta$   \\
%			& \multicolumn{3}{c}{\bf Cross} & \multicolumn{3}{c}{\bf BoF} & \multicolumn{3}{c}{\bf SS-BoF} \\
%\hline
% 	& 50\% & 30\% & 10\%  	& 50\% & 30\% & 10\% &  50\% & 30\% & 10\% \\
%\hline
%
%{\bf $L_2$ BoF}	 	& & & & 96.97\% & 96.97\% & 96.97\% & & &     \\
%{\bf $L_2$ SS-BoF}	 & & & & & & & 95.29\%  & 95.29\%   & 95.29\%      \\
%\hline
%{\bf MM-NN}     & {\bf 99.53}\% & {\bf 99.17}\% & {\bf 99.28}\%  	   & {\bf 99.67}\% & {\bf 99.63}\%  & {\bf 99.63}\%     & {\bf 99.53}\% & {\bf 99.17}\% & {\bf 99.26}\% \\
%{\bf CM-NN}     & 98.44\%  & 96.91\% & 95.03\% &  98.96\% & 97.59\% & 94.35\%     & 98.51\% & 98.40\% & 95.83\%   \\
%\hline
%{\bf CM-SSH}	      &  - to be - \%  added    & \%     &  \%   & & & & & & \\	
%\hline
%\end{tabular}
%\end{center}\vspace{-3mm}
%\caption{\label{tab:shapegoogle-p-at-k-missingdata} Performance of different methods (mAP) on the ShapeGoogle retrieval experiment where only a subset of the cross-modal correspondences is kept . Hash of length $m=16$ is used and the best configuration is found through grid search.}
%\end{table}

\begin{figure*}[!h]
\begin{center}
\includegraphics[width=\linewidth]{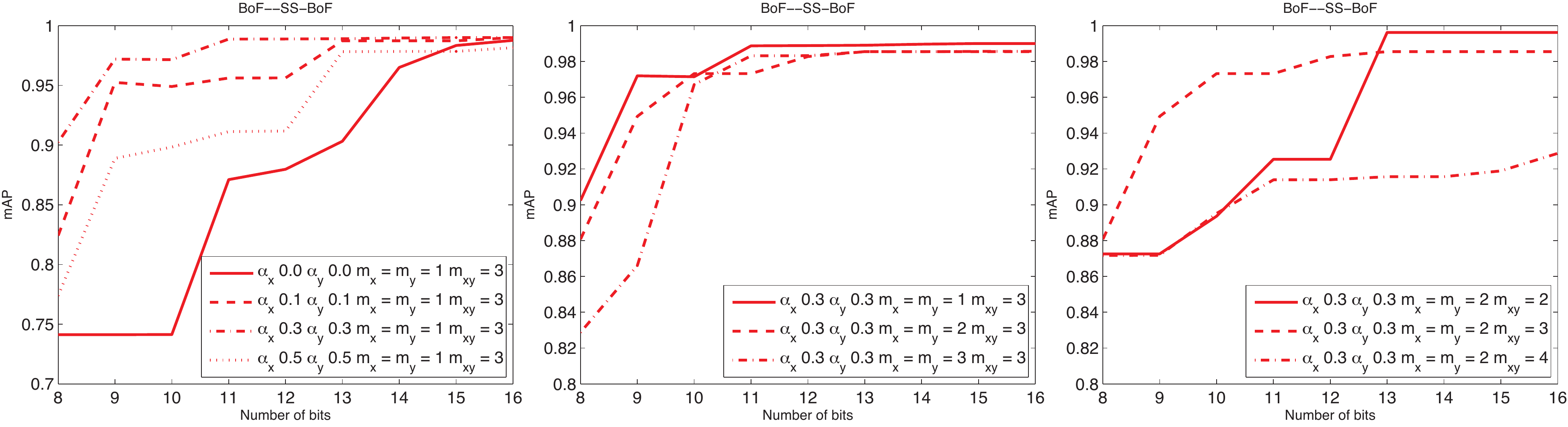}
\end{center}
\caption{mAP for the ShapeGoogle experiment with various configuration of the hyper-parameters of the system.}
\label{fig:retrieval-shapegoogle-gridsearch}
\end{figure*}

\paragraph{Choice of parameters}  
The theoretically smallest hash length must be $m = \lceil \log_2(\#classes) \rceil$.  % $m \geq \log_2(\#classes)$. 
However, since we are using a simple embedding, in practice an $m$ about 5--10 times larger may be required to achieve satisfactory results. 
Table~\ref{tab:shapegoogle-params} and  Figure~\ref{fig:retrieval-shapegoogle-gridsearch} show the performance of the NNhash methods under different choices of the parameters.
%MM-NN is less sensitive than CM-NN to the choice of the parameters and all results are close to perfect accuracy. 
We can see that the addition of intra-modal regularization makes the cross-modal performance less sensitive to the choice of the parameters, and that MM-NN produces higher cross-modal performance than CM-MM for similar margin settings.

%%%%%%%%%%%%%%

\begin{table}[htdp]
\begin{center}
\begin{tabular}{rccccccccccc}
%			&  \bf{ p@1} 	& \bf{ p@10}  \\
& & & &  & \multicolumn{2}{c}{ \hspace{-1mm} {\bf BoF--SS-BoF}  \hspace{-1mm}  } & \multicolumn{2}{c}{ \hspace{-1mm} {\bf BoF}  \hspace{-1mm}  } & \multicolumn{2}{c}{ \hspace{-1mm} {\bf SS-BoF}  \hspace{-1mm}  } \\
		$\hspace{-1mm}\alpha_X$\hspace{-1mm} & \hspace{-1mm}$\alpha_Y$\hspace{-1mm} & \hspace{-1mm}$m_X$\hspace{-1mm} & \hspace{-1mm}$m_Y$\hspace{-1mm} & \hspace{-2mm} $m_{XY}$\hspace{-2mm} 
& 12 bit & 16 bit & 12 bit & 16 bit & 12 bit & 16 bit\\
\hline
%0.0  & 0.1   &  1 & 1 & 3									         & 98.83\%    \\
%0.0  & 0.3   &  & &  										         & 98.36\%    \\
0.0  & 0.5   & 1 & 1 & 2										& 90.67\% & 99.38\%    & 91.26\% & 98.87\%   & 90.67\% & 99.38\% \\
0.5  & 0.0   & 2 & 2 & 3										& 93.12\% & 99.39\%    & 94.83\% & 99.35\%   & 91.79\% & 99.25\%  \\
%0.1  & 0.0   &  & &  											& 98.98\%    \\
%0.3  & 0.0   &  & &											& 98.59\%    \\ 
%0.0  & 0.3   &  & &											& 98.63\%    \\ 
0.1  & 0.1   & 1 & 1 & 3										& 95.62\% & 99.00\%    & 96.05\% & 98.93\%   & 95.62\%  & 99.00\%  \\ 
%0.1  & 0.3   &  & &											& 99.69\%    \\ 
%0.1  & 0.5   &  & &											& 97.39\%    \\ 
%0.3  & 0.1   &  & &											& 98.43\%    \\ 
%0.5  & 0.1   &  & &											& 97.97\%    \\ 
0.3  & 0.3   & 2 & 2 & 2										& 92.54\% & 99.62\%    & 92.51\% & 99.45\%  & 92.54\% & 99.62\%\\ 
0.3  & 0.3   & 1 & 1 & 3										& 98.88\% & 99.01\%    & 98.97\%  & 99.24\% & 98.88\% & 99.01\% \\ 
0.3  & 0.3   & 2 & 2 & 3										& 98.28\% & 98.55\%    & 98.74\%  & 98.51\% & 98.28\% & 99.48\% \\ 
%
%0.3  & 0.5   &  & &											& 98.11\%    \\ 
%0.5  & 0.3   &  & &											& 98.63\%    \\ 
0.5  & 0.5   & 3 & 3 & 3										& 95.59\% & 99.07\%    & 95.46\% & 99.13\% & 95.59\% & 98.93\% \\ 
\hline 
 \multirow{3}{*}{0}  &  \multirow{3}{*}{0} & \multirow{3}{*}{0} & \multirow{3}{*}{0} & 2      & 96.03\% & 96.96\% & 96.52\% & 96.97\% & 95.51\% & 96.72\% \\ 
			     & 				     & 				   &  				 & 3	   & 87.97\% & 98.77\%  & 88.60\%& 99.64\% & 87.97\% & 98.99\% \\
			     & 				     &      			   &  				 & 5 	   & 91.22\% & 95.60\%  & 93.75\%& 97.06\% & 91.24\% & 95.60\% \\ 
\hline
\end{tabular}
\end{center}\vspace{-3mm}
\caption{\label{tab:shapegoogle-params} Performance of different methods in the ShapeGoogle cross-modal retrieval experiment using several hash lengths and various selection of the parameters for single layer nets. The settings with $\alpha_{X} = \alpha_{Y} = m_X = m_Y = 0$ correspond to CM-NN.
%Hashes of 12-bit are produced cropping the 16-dim codes and hence they are usually sub-optimal (the problem is not separable).
} %For MM-NN we used $m_{XY} = 49$ and $m_X = m_Y = 9$.}
\end{table}

%%%%%%%%%%%%%%

%\begin{table}[htdp]
%\begin{center}
%\begin{tabular}{rccc}
%%& \multicolumn{3}{c}{\bf mAP}  \\
%%\hline
%%			&  $\xi \times \eta$ & $\xi \times \xi$ & $\eta \times \eta$   \\
%			& {\bf Cross} & {\bf BoF} & {\bf SS-BoF} \\
%\hline
%{\bf $L_2$ BoF}	 	& & 96.97\% &    \\
%{\bf $L_2$ SS-BoF}	 & & & 95.29\%      \\
%\hline
%{\bf MM-NN}     & \bf{98.23}\%  & 98.24\% & \bf{98.22}\%  \\
%{\bf CM-NN}     & 97.86\%         & \bf{98.26}\% & 97.82\% \\
%\hline
%{\bf CM-SSH}	      &  72.37\%         & 82.95\%     &  96.42\%   \\	
%\hline
%\end{tabular}
%\end{center}\vspace{-3mm}
%\caption{\label{tab:shapegoogle-p-at-k} Performance of different methods (mAP) on the ShapeGoogle retrieval experiment. Hash of length $m=32$ is used.}
%\end{table}

%\subsection{NUS-WIDE}
\paragraph{NUS} 
In the second experiment, we used the NUS dataset of \cite{nus-wide-civr09}, containing about 250K annotated images from Flickr. 
The images are manually categorized into 81 classes (one image can belong to more than a single class) and represented as 500-dimensional bags of SIFT features (BoF, used as the first modality) and 1000-dimensional bags of text tags (Tags, used as the second modality). 
%
%The latter is a web image dataset which contains $\approx$250k images with the corresponding Flickr tags. Six types of features are available for each image including 64-D colour histogram, 144-D colour correlogram, 73-D edge direction histogram, 128-D wavelet texture, 225-D block-wise colour moments and 500-D bag of words based on SIFT descriptions.
%We select the 500 dimensional bag of words (Bofs) representation for modality $1$ and the binary vector of Tags (1k dimensional, a dimension for each tag) for second modality.
%
%
The dataset was split into approximately equal parts for testing and training. 
We used positive and negative sets of size $|\mathcal{P}| = |\mathcal{N}| = 5\times 10^5$. 
Positive pairs were images belonging at least to one common class; negative pairs were images belonging to disjoint sets of classes. 
For MM-NN, we used the margins  $m_{XY} = 7, m_X = m_Y = 3$ and $\alpha_{X} = \alpha_{Y} = 0.3$.  CM-NN used $m_{XY} = 7$.
Testing was performed using a query and database sets of size approximately $10^4$ and $1.8\times 10^5$, respectively. First ten matches were found using approximate nearest neighbors \cite{ann}. 
Matches that had at least one class in common with the query were considered correct. 

\begin{figure*}[!htb]
\begin{center}
\includegraphics[width=\linewidth]{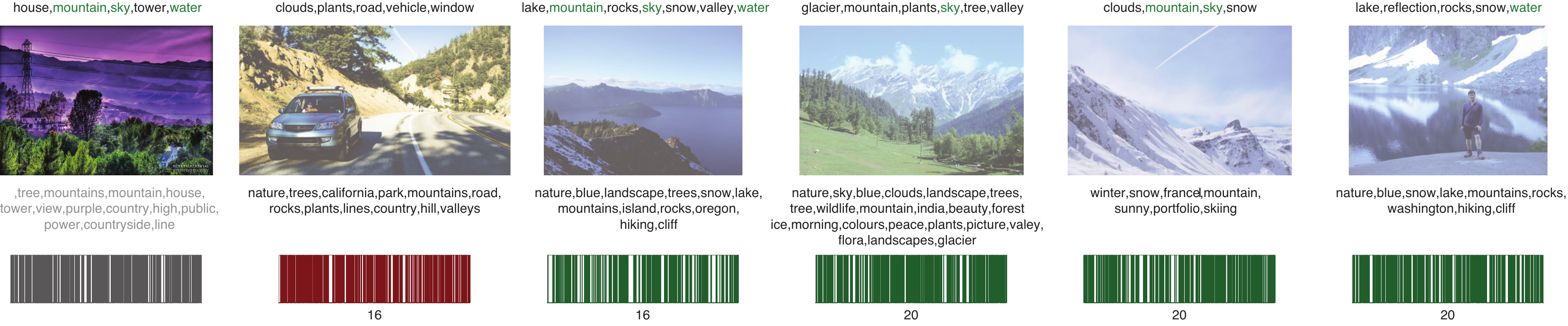}\vspace{2mm}\\
\includegraphics[width=\linewidth]{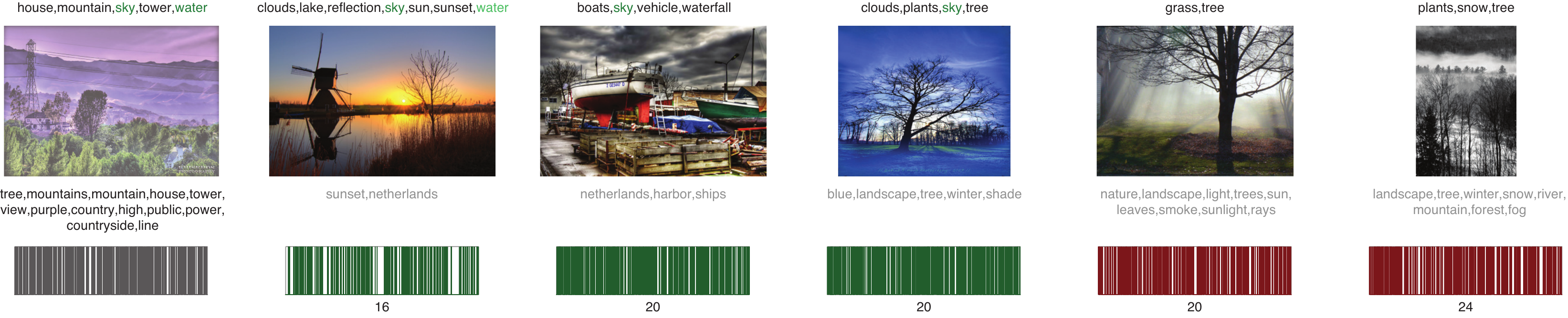}\vspace{5mm}\\
\includegraphics[width=\linewidth]{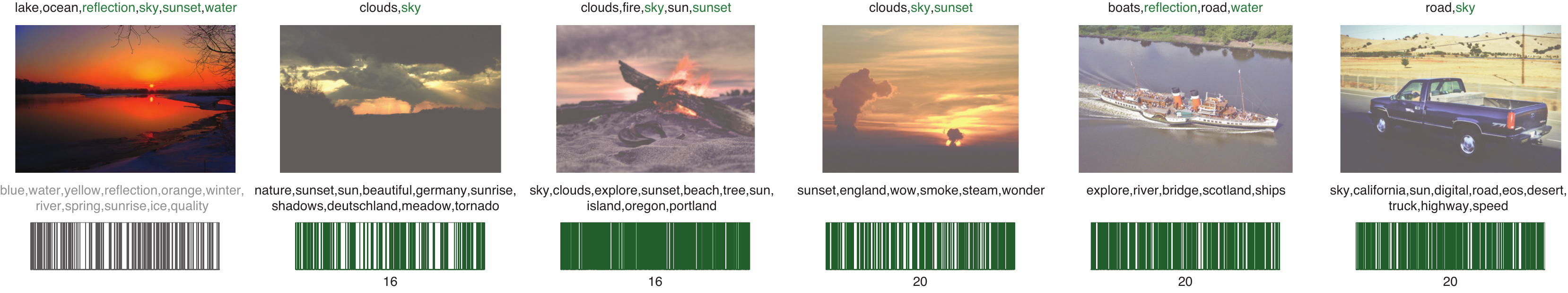}\vspace{2mm}\\
\includegraphics[width=\linewidth]{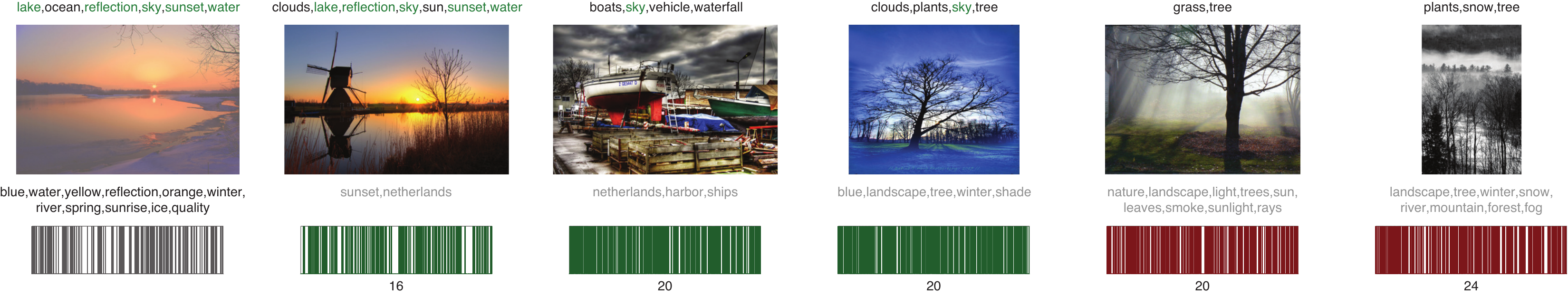}\vspace{-5mm}\\
\end{center}
\caption{Top 5 matches for MM-NNhash in BoF--Tags (rows 1-5) and Tags--BoF (rows 6-10) cross-modal retrieval. 
Rows 1, 6: classes; rows 2, 7: images; rows 3, 8: tags (subdued implies the modality was not used); rows 4, 9: resulting hashes; rows 5, 10: Hamming distance from query. 
%Correct matches are shown in green. Classes are shown above the images; tags are shown below the images.
}
\label{fig:retrieval-nus}
\end{figure*}

%Figure~\ref{fig:nus} and 
Table~\ref{tab:nus-map-table} compares the performance of different methods. MM-NN outperforms other approaches in all quality criteria. Figure~\ref{fig:retrieval-nus} shows examples of top matches using MM-NN. 
%Concepts are shown above the images whereas the corresponding labels and hash codes are shown below.
%

%%%%%%%%%%%%%%%%%%%%%%%%%%%%%%%%%%%%%%%%%%%%%%%%%
% For CM-SSH exp loss and alpha = 0.2 were used. A fresh train has been performed after 
% the previous ICML submission.
% For L2 results are with cutoff at 10 as all the others. Double checked.
%%%%%%%%%%%%%%%%%%%%%%%%%%%%%%%%%%%%%%%%%%%%%%%%%
\begin{table}[htbp]
\begin{center}
\begin{tabular}{rrcccc}
			&& {\bf BoF--Tags} & {\bf Tags--BoF} & {\bf BoF} & {\bf Tags} \\
\hline
%\multicolumn{2}{r}{\bf Raw} 	 &	-- & -- & 57.06\% & 78.87\%   \\
\multicolumn{2}{r}{\bf Raw} 	 &	-- & -- & 57.0\% & 78.9\%   \\
\hline
 \multicolumn{2}{r}{\bf MM-NN }  & {\bf 64.3\%}  &  {\bf 55.00\%} &  {\bf 79.39\%} &  {\bf 87.41\%}  \\
 \multicolumn{2}{r}{\bf CM-NN }   & 61.1\%    &  51.3\%   &   75.62\% &  86.23\% \\
 
 %%
 % These are results from 64-0301-500e-normBT-00
 %
% \multicolumn{2}{r}{\bf MM-NN }  & {\bf 60.0\%}  &  {\bf 62.3\%} &  {\bf 60.2\%} &  {\bf 77.0\%}  \\
% \multicolumn{2}{r}{\bf CM-NN }   & 58.8\%    &  55.0\%   &   59.7\% &  74.5\% \\ 
 %%
 
% \multirow{2}{*}{\bf MM-NN }  &  {\bf L1} & 60.01\%  &  62.27\% &  79.39\% &  87.41\%  \\
% & {\bf L2}   &  \%  & \% & \% &  \\
% \multirow{2}{*}{\bf CM-NN } & {\bf L1}    & 58.75\%    &  54.96\%   &   75.62\% &  86.23\% \\
% & {\bf  L2}  & --\%         & --\% & --\% & \\
\hline
%\multicolumn{2}{r}{\bf CM-SSH}	      &  53.73\%    &  50.24\%  & 54.09\%     &  75.99\%   \\	
\multicolumn{2}{r}{\bf CM-SSH}	      &  53.7\%    &  50.2\%  & 54.0\%     &  76.0\%   \\	
\hline
\end{tabular}
\end{center}\vspace{-3mm}
\caption{\label{tab:nus-map-table} mAP of different methods for the NUS experiment. Hashes of length $64$ were used.}
\end{table}
%%%%%%%%%%%%%%%%%%%%%%%%%%%%%%%%%%%%%%%%%%%%%%%%%

\begin{figure*}[!ht]
\begin{center}
%\includegraphics[width=\linewidth]{figures/NUS_animal.png}\\
%{\small animal}\vspace{3mm}\\
%
%\includegraphics[width=\linewidth]{figs/NUS_cloud.png}\\
\includegraphics[width=0.48\linewidth]{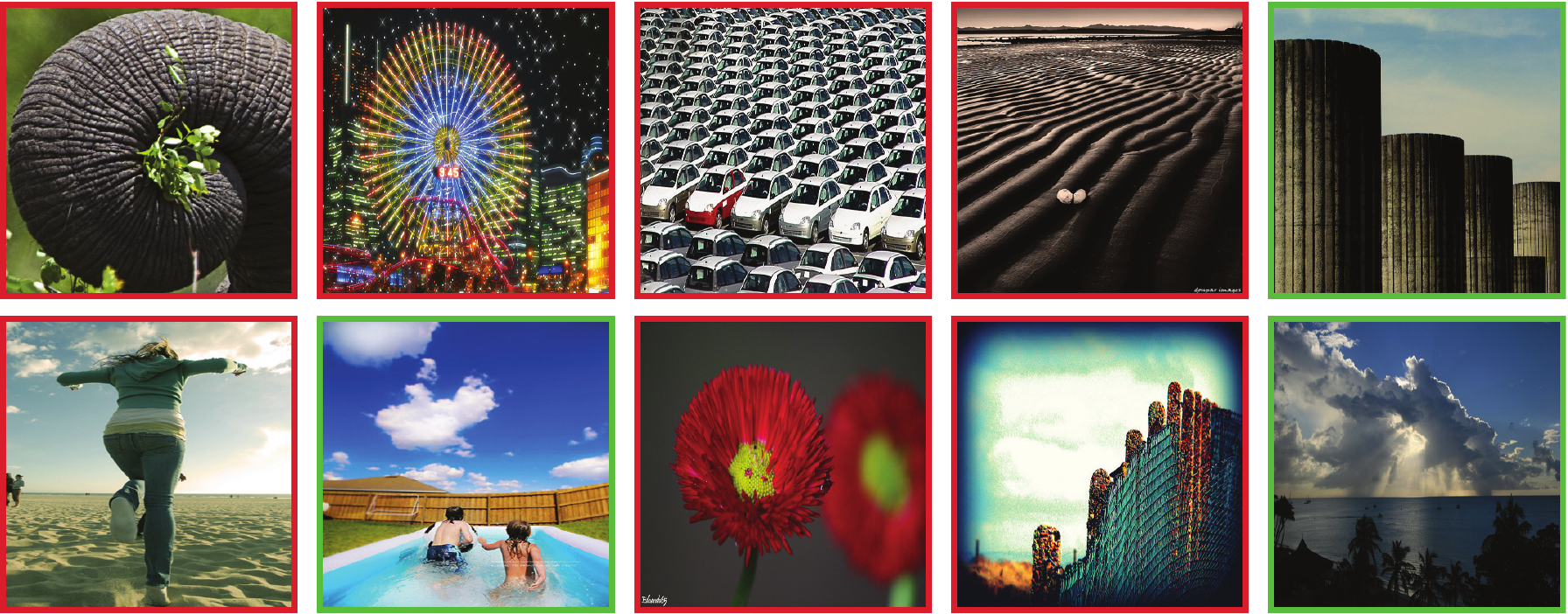}
%{\small cloud}\vspace{3mm}
%
\hspace{2mm}
\includegraphics[width=0.48\linewidth]{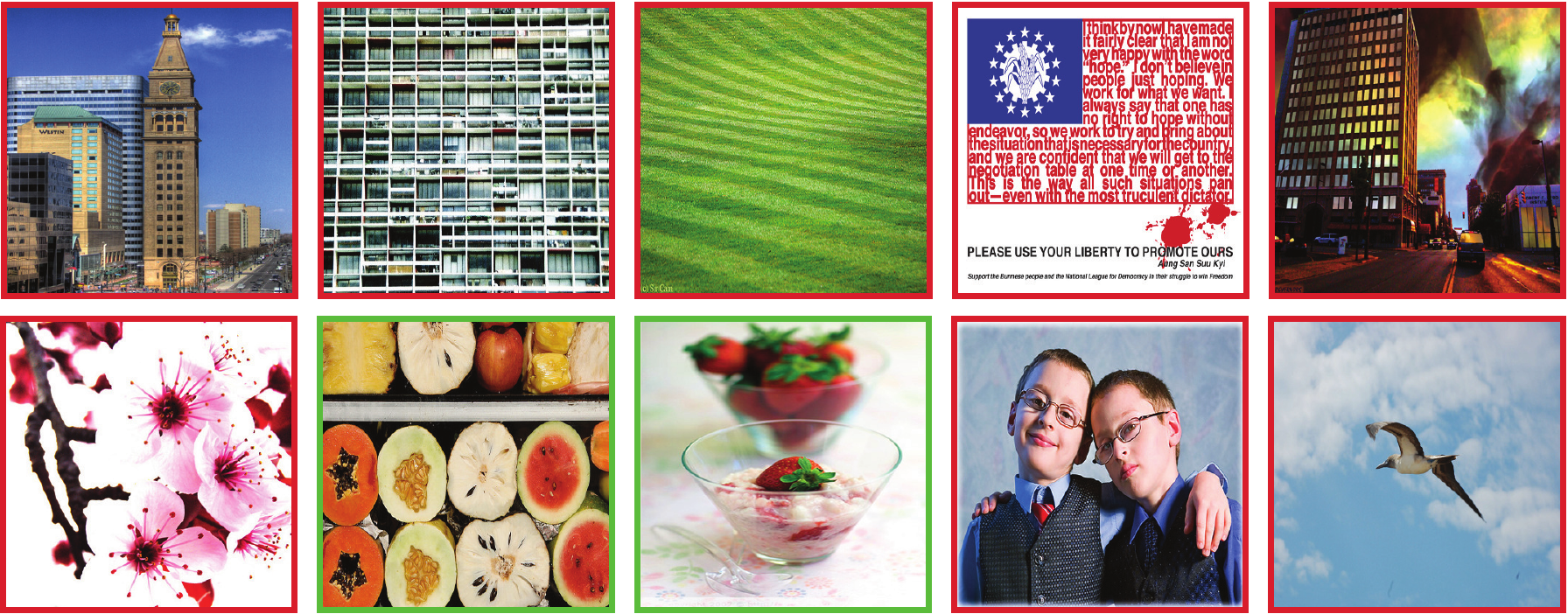}\\
\hspace{0cm}{\small cloud}\hspace{6cm} {\small red+food}\vspace{-3mm}\\
\end{center}
\caption{Example of text-based image retrieval on NUS dataset using multimodal hashing. Shown are top five image matches produced by CM-SSH (top) and MM-NN (bottom) in response to two different queries: {\em cloud} (left) and {\em red+food} (right). 
Relevant matches %(having at least one class in common with query) 
are shown in green.}
%Hashes are matched to BoFs to evaluate cross-modal retrieval.}
\label{fig:retrieval-nus-tags-bofs}
\end{figure*}

\begin{figure*}[!ht]
\begin{center}
%\includegraphics[width=\linewidth]{figures/NUS_animal.png}\\
%{\small animal}\vspace{3mm}\\
%
%\includegraphics[width=\linewidth]{figs/NUS_cloud.png}\\
\includegraphics[width=0.48\linewidth]{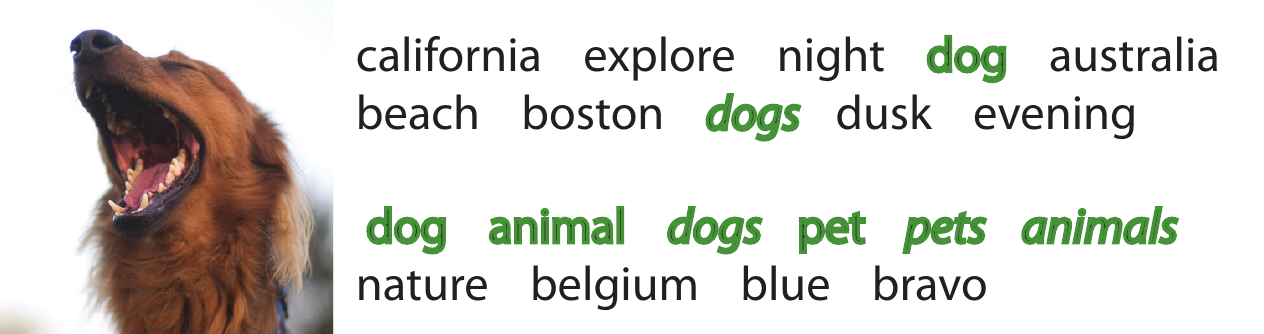}
%{\small cloud}\vspace{3mm}
%
\hspace{3mm}
\includegraphics[width=0.48\linewidth]{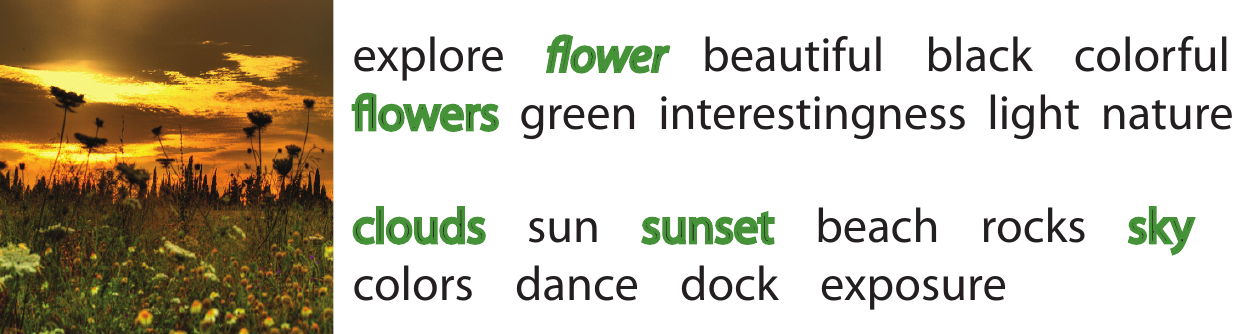}\vspace{-3mm}
\end{center}
\caption{Example of image annotation on NUS dataset using multimodal hashing. Shown are tags returned for image query using CM-SSH (top) and MM-NN (bottom). Groundtruth tags are shown in green; synonyms are italicized.}
%Hashes are matched to BoFs to evaluate cross-modal retrieval.}
\label{fig:retrieval-nus-bofs-tags}
\end{figure*}

Figure~\ref{fig:retrieval-nus-tags-bofs} shows retrieval results using as queries artificially created Tag vectors containing specific words such as ``cloud''. These tags are hashed using $\eta$ and matched to BoFs hashed using $\xi$. The retrieved results are meaningful and most of them belong to the same class. It is especially interesting to note that MM-NN, apart from the Labels in the ground-truth which are in general noisy, produced relevant results in both cases. %; e.g. the first images clearly contain, respectively a cloud and a red colored object.
Figure~\ref{fig:retrieval-nus-bofs-tags} shows image annotation results. We retrieve the top five Tags matches from a BoF query and assign the ten most frequent annotations to the image. We clearly see that MM-NN produces better annotations than CM-SSH also in this case.

%The dataset is generated using 500k positive and negative couples and we considered as positive every couple with non empty concepts' intersection.
%For our NNhash models we normalise the input vectors as follows. For Bofs we normalise each pattern to have unit length whereas for Tags we simply remap $0$ elements to $-1$.

%\subsection{Cross-Modal Retrieval}
%To further validate the performance of our hashing method,  we test it on a cross-modal image retrieval task. 
%We create a database of hashes for Bofs modality using the ANN algorithm \cite{ann} and we query it using new Tags hashes; that is we generate a vector where only selected elements are set.
%We decide to adopt an approximate nearest neighbour approach as querying a db of K elements with 10K queries would be otherwise prohibitive.
%In fig.\ref{fig:retrieval-nus} top 5 images returned by various query for the cross-modal Tags to Bofs retrieval are shown.

%%%%

\paragraph{Wiki} In the third experiment, we reproduced the results of \citet{rasiwasia2010new} using the dataset of 2866 annotated images from Wikipedia. 
The images are categorized in 10 classes and represented as 128-dimensional bags of SIFT features (Image modality) and 10-dimensional LDA topic model (Text modality).  
The dataset was split into disjoint subsets of 2173 and 693 for training and testing, respectively.  
We used positive and negative sets of size $|\mathcal{P}| = 1\times 10^4$, $|\mathcal{N}| = 1\times 10^5$. 
Table~\ref{tab:wiki-map} %and Figure~\ref{fig:retrieval-wiki} 
shows the mAP for the Image--Text and Text--Image cross-modal retrieval experiment. 
For reference, we also reproduce the results reported in \cite{rasiwasia2010new} using correlation matching (CM), semantic matching (SM), and semantic correlation matching (SCM). MM-NN slightly outperforms SCM on average. 
%in Image-Text queries and works slightly worse in Text-Image retrieval. 
%is about $27.7\%$ mAP for Image query and $22.6\%$ for Text query. 
%
We should stress however that these results are not directly comparable with ours: while \cite{rasiwasia2010new} find a Euclidean embedding, we use Hamming embedding (in general, a more difficult problem). While having similar performance to SCM, the significant advantage of our approach is that it produces much smaller compact binary codes (at least 10$\times$ smaller) that can be searched very efficiently.

%
%
%For MM-NNhash, we used the margins  $m_{XY} = 49, m_X = m_Y = 9$ and $\alpha_{X} = \alpha_{Y} = 0.3$.  
%
%
%Testing was performed using a query and database sets of size approximately $10^4$ and $\approx1.8\times 10^5$, respectively. First 10 matches were found using approximate nearest neighbors \cite{ann}. 
%
%Matches that had at least one class in common with the query were considered correct. 

\begin{table}[htdp]
\begin{center}
\begin{tabular}{rrccccc}
%			&  \bf{ p@1} 	& \bf{ p@10}  \\
			&& {\bf Image-Text } & {\bf Text-Image } & {\bf Average } \\
\hline
 \multirow{2}{*}{\bf MM-NN}    & {\bf L1}  &  27.8\%   &  21.2\%   & 24.5\%\\  
  & {\bf L2}   &  {\bf 28.5\%}   & 22.0\% & {\bf 25.3\%}\\  
  \hline
 \multirow{2}{*}{\bf CM-NN }   & {\bf L1}    &  26.7\%   &  20.9\% & 23.8\%\\  % these results are on clusy
    & {\bf L2}   &  27.1\%  & 21.1\% & 24.1\% \\  % these results are on clusy
\hline
\multicolumn{2}{r}{ {\bf CM-SSH} } & 22.2\%  & 18.4\%  & 20.3\% \\  
\hline
\multicolumn{2}{r}{ {\bf CM}$^*$ } & 24.9\% & 19.6\%  & 22.3\% \\
\multicolumn{2}{r}{ {\bf SM}$^*$ } & 22.5\% & 22.3\%  & 22.4\% \\
\multicolumn{2}{r}{ {\bf SCM}$^*$ }& 27.7\% & {\bf 22.6\%}  & 25.2\% \\
\hline
\end{tabular}
\end{center}\vspace{-3mm}
\caption{\label{tab:wiki-map} Performance of different methods in the cross-modal retrieval experiments (Image and Text modalities) 
%(BoF to Tags matching) 
on the Wiki dataset. 32-bit hashes were used. Results marked with * are from \cite{rasiwasia2010new} based on Euclidean embedding; these results are not directly comparable with hashing and are brought here for reference only. }
\end{table}

\section{Conclusions}

We introduced a novel learning framework for multimodal similarity-preserving hashing based on the coupled siamese neural network architecture.
Our approach is free from assuming linear projections unlike existing cross-modal similarity learning methods; in fact, by increasing the number of layers in the network, mappings of arbitrary complexity can be trained (our experiments showed that using multilayer architecture results in a significant improvement of performance). 
We also solve the exact optimization problem during training making no approximations like the boosting-based CM-SSH. Our method does not involve semidefinite programming, and is scalable to a very large number of dimensions and training samples.
%To the best of our knowledge, this is also the first attempt to combine intra- and inter-modality similarity learning.
%
Experimental results on standard multimedia retrieval datasets showed performance superior to state-of-the-art hashing approaches.

%The proposed MM-NN hash and its cross-modal CM-NN variant outperform the state-of-the-art CM-SSH algorithm in the ShapeGoogle and NUS multi-media retrieval tasks. 

%
%
%\begin{figure}[ht]
%\begin{center}
%\includegraphics[width=.9\linewidth]{figs/mAP_I_T.eps}\\
%\end{center}\vspace{-5mm}
%\caption{Performance of different methods (mAP) on the Wiki retrieval experiment. Hash of length $m=32$ is used. }
%%Hashes are matched to BoFs to evaluate cross-modal retrieval.}
%\label{fig:retrieval-wiki}
%\end{figure}
%
%

%\clearpage
%\small
\setlength{\bibspacing}{0.2\baselineskip}
\bibliography{mmhash}
\bibliographystyle{icml2012}

\end{document}